%% file: 0.main.tex
\documentclass{cvmp_short}
\usepackage{microtype}
\usepackage{float}
\usepackage{wrapfig}
\usepackage{amsmath}
\usepackage{amsbsy}
\usepackage{amsfonts}
\begin{document}
\title{Face Reflectance and Geometry Modeling via Differentiable Ray Tracing}
\addauthor{Abdallah Dib$^1$, Gaurav Bharaj$^2$, Junghyun Ahn$^2$, \\C\'edric Th\'ebault$^1$, Philippe-Henri Gosselin$^1$, Louis Chevallier}{}{1}
\addinstitution{
InterDigital, Rennes, France}
\addinstitution{
 Technicolor, New York, NY}
\maketitle
%
%
\noindent
\vspace{-35px}
\input{1.introduction}
\vspace{-10px}
\input{2.facemodeling}
\vspace{-10px}
\input{3.optimization}
\input{4.results}
\bibliography{0.main}
\bibliographystyle{ieeetr}
\end{document}

%% file: 1.introduction.tex
\paragraph{Introduction.} A classic problem in computer vision is to reconstruct 3D faces from monocular image and video data. This needs disentangle of scene and face parameters such as light, facial geometry (pose, identity, and expression), and reflectance, while accounting for camera depth ambiguity. Image pixel saturation due to self-shadows and specularity further add to the problem complexity.
\\
To model the problem, optimization based~\cite{garrido2016reconstruction} and more recently deep learning based~\cite{tewari2019fml} methods have been proposed. A photo-consistency loss difference between a parametric 3D face render and input facial image helps estimate the scene and face parameters. Often the optimization cost is regularized with statistical priors and mathematical models of the parameters. \cite{yamaguchi2018high} train a deep neural net to reconstruct faces and rely on high-quality training data obtained via a photometric camera rig. Both \cite{yamaguchi2018high} and \cite{gotardo2018practical} present reflectance models with diffuse and specular albedo. They do not handle self-shadows explicitly.
\\
We present a novel method to model 3D faces from monocular images, with parameterized ray traced scene light(s), and explicit disentangle of facial geometry (pose, identity and expression), reflectance (diffuse and specular albedo), and self-shadows (see Figure~\ref{fig:teaser}). We take inspiration from \cite{debevec2000acquiring} and model the scene light as a {\it virtual light stage} with pre-oriented area lights. This setup is used in conjunction with a differentiable Monte-Carlo ray tracer~\cite{li2018differentiable}, to optimize the scene and face parameters. Due to the disentangle of the parameters, we can not only obtain robust results, but also gain explicit control over them, with several practical applications. For example, we can change facial expression with accurate resultant self-shadows or relight the scene and obtain accurate specular reflection.

%% file: 2.facemodeling.tex
\begin{wrapfigure}{L}{0.15\textwidth}
\vspace{-10px}
\includegraphics[width=0.15\textwidth]{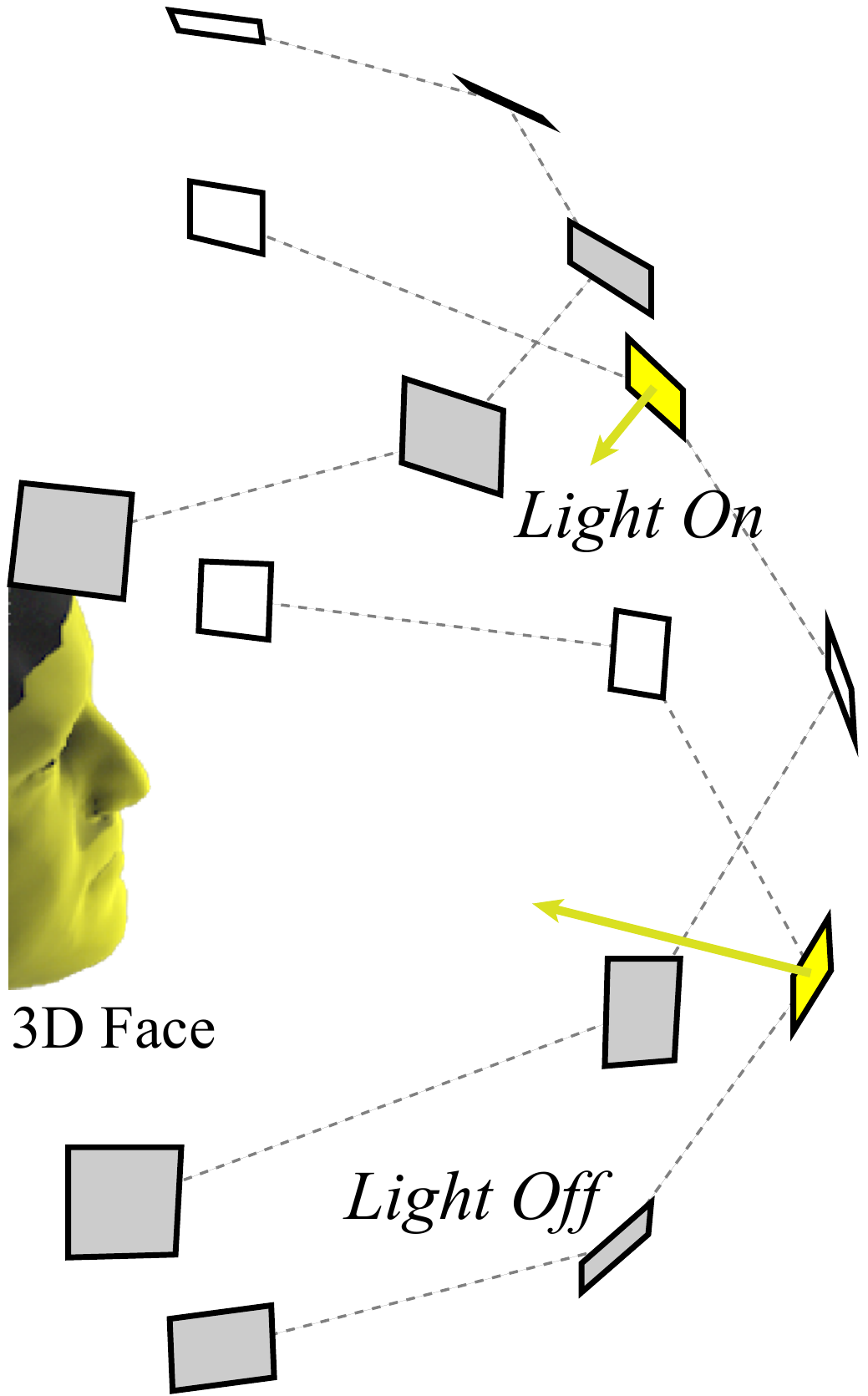}
\vspace{-10px}
\caption{A \textit{virtual light stage}}
\label{fig:lightstage}
\vspace{-10px}
\end{wrapfigure}
\paragraph{Face Modeling.} Similar to \cite{garrido2016reconstruction}, our method follows an optimization-based face reconstruction methodology. \cite{garrido2016reconstruction,tewari2019fml} model scene light via spherical-harmonics, and reflectance via the Lambertian (diffuse albedo only) model. Consequently, they cannot handle self-shadows and specularities. Image pixels with saturated color values due to self-shadows or specularity can lead to an underconstrained photo-consistency loss based optimization. Here, face geometry vertices corresponding to saturated pixel color values can deform the geometry unnaturally. We model reflectance via the Cook-Torrance BRDF (diffuse and specular albedo) \cite{pharr2016physically}, and use physically accurate ray tracing to handle parameterized area lights and resultant self-shadows.
\\
In order to render physically accurate images via ray tracing, scene geometry and light must be known. To this end, we create a parameterized virtual light stage with $N$ fixed oriented (pointing towards origin) area lights placed on vertex positions of an isotropically sampled hemisphere (see Figure~\ref{fig:lightstage}). Parameterized face geometry is placed at the light stage's origin, facing the camera. We then employ \cite{li2018differentiable}'s differentiable ray tracing method to render an image. Finally, during face reconstruction optimization, backpropagation automatically calculates the gradient for the various parameters w.r.t photo-consistency cost loss between the input and rendered images.
\\
The scene light parameters are the {\it color} intensities $\{\mathbf{l}_i\} \in N(=20) \text{ and } \mathbf{l}_i \in \mathbb{R}^3$. A \textit{light-off} (Fig~\ref{fig:lightstage}) implies intensity $\mathbf{0}^T$. The face geometry parameters are $\alpha \in \mathbb{R}^{80}$ which control a 3D model morphable based facial identity, $\beta \in \mathbb{R}^{90}$ controls facial blend shape expressions, and $[\mathbf{R}\;\;\mathbf{t}] \in \mathbb{R}^{6}$ the geometry transform. The skin reflectance model is given by Cook-Torrance reflectance $f_r = k_d f_{lambertian}(\mathbf{l}_j, \delta, \mathbf{n}_i) + k_s f_{cook−torrance}(\mathbf{l}_j, \mathbf{c}_{spec}, \mathbf{n}_i)$. Here, $\delta \in \mathbb{R}^{80}$ are the morphable model's vertex color albedo, $\mathbf{n}_i(\alpha, \beta, \mathbf{R}, \mathbf{t})$ the vertex normal and $\mathbf{c}_{spec} \in \mathbb{R}^{32\times32}$ are the pixel values of a vectorized gray-scale specular texture map.

%% file: 3.optimization.tex
\paragraph{Optimization.} The photo-consistency optimization is given by:
\vspace{-7px}
\begin{equation}
\vspace{-7px}
    \operatorname*{argmin}\;\;E_{data} + w_1 E_{prior} +  w_2 E_{light}, \label{eq:main}
\end{equation}
where $E_{data}=\sum_{i\in\mathcal{I}}||\mathbf{p}_i^T-\mathbf{p}_i^R||_2^2$ and $E_{light}=\sum_{j=0}^N||\mathbf{l}_j-\mathbf{m}_j||_2^2$. The vectors $\mathbf{p}_i^T, \mathbf{p}_i^R$ are ray traced and real image ($\mathcal{I}$) pixel color values, respectively and $\mathbf{m}_j$ is the mean intensity of the $j^{th}$ light. In order to make the optimization tractable, we use landmark and statistical prior $E_{prior}$ defined in \cite{garrido2016reconstruction}. Our method relies on geometry, reflectance, and light parameters to calculate the final render pixel color values $\mathbf{p}_i^T=\mathcal{F}(f_r, \mathbf{l}_j, \alpha, \beta, \mathbf{R}, \mathbf{t})$, where $\mathcal{F}$ is the Monte-Carlo estimator of the lighting equation. Generally, it is desirable to minimize light flux, to this end, we add a light flux conservation regularizer $E_{light}$ to Equation \ref{eq:main}.
\begin{figure}[t]
\vspace{-10px}
  \includegraphics[width=\linewidth]{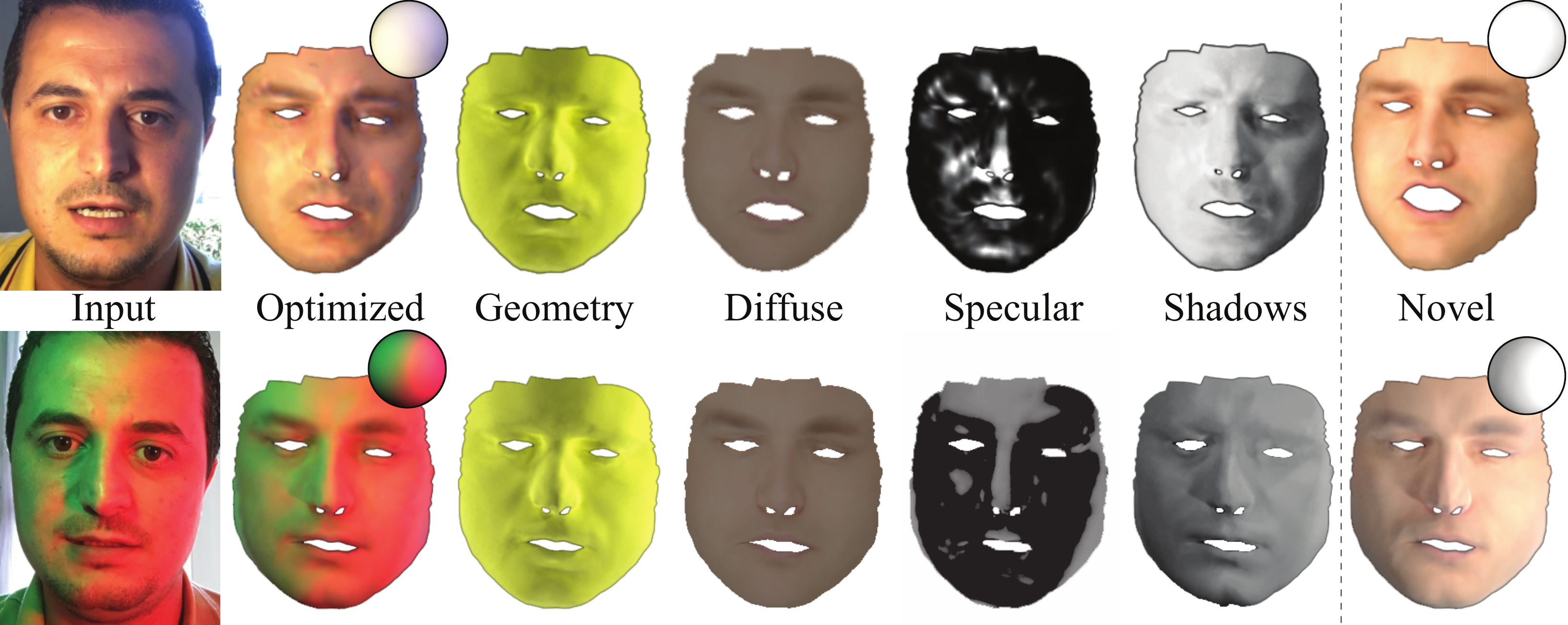}
  \vspace{-15px}
  \caption{(From left to rigt) Given an input image, we optimize the scene light (shown via spherical environmental map), geometry (identity and expression), reflectance (diffuse and specular albedo) parameters with self-shadows. Top: We optimize for uniform color light(s). The last column shows updated shadows, as we transform the geometry and change the facial expression. Bottom: We optimize for multi color lights. The last column shows a relit render with accurate reflectance and shadows.}
  \label{fig:teaser}
\vspace{-10px}
\end{figure}
\\
We optimize the parameter in a sequential strategy as described in \cite{garrido2016reconstruction}'s Algorithm 1 - Multi-Step Optimization Strategy. In addition, we add the following steps to the optimization sequence: (1) Cook-Torrance reflectance based specularity, and (2) light ($\mathbf{l}_j$) optimization. The `shadowed' regions of the 3D face geometry are calculated by the ray tracer automatically.

%% file: 4.results.tex
\vspace{-10px}
\paragraph{Results and Conclusion.} In Figure~\ref{fig:teaser} (Top), we show face reconstruction with disentangled reflectance (diffuse and specular albedo), light (visualized as a spherical environment map), geometry (identity and expression), and self-shadow, especially near the nose region. While, Figure \ref{fig:teaser} (Bottom) shows accurate self-shadow based on optimized multi-color lights. Notice how the change in scene light (last column) results in correct updated reflectance and self-shadows. Disentangled light, reflectance and shadows result in accurate reconstruction of face geometry regions that correspond to image patches with saturated pixel colors (due to shadows or specularities) and does not deform the geometry unnaturally.
\\
We present a novel strategy to automatically reconstruct faces with ray traced light, specularities, and shadows. The method robustly handles scenes with strong directional lights, partial occlusion due to shadows and specularities. By using a differentiable ray tracer, we accurately optimize the photo-consistency loss. 
\\
However, our method is limited by the ray tracer's sampling rate. A low sampling rate results in a noisy render, while higher sampling rate results in slow optimization convergence. Another limitation is that our reflectance model does not model skin's subsurface scattering properties.